# Peanut Maturity Classification using Hyperspectral Imagery

Sheng Zou[a*], Yu-Chien Tseng[b, c*], Alina Zare[a], Diane Rowland[b], Barry Tillman[b,d], Seung-Chul Yoon[e]

[a] Department of Electrical and Computer Engineering, University of Florida, Gainesville, Florida 32611, USA
[b] Department of Agronomy, University of Florida, Gainesville, Florida 32611, USA
[c] Department of Agronomy, National Chiayi University, Chiayi 60004, Taiwan
[d] North Florida Research and Education Center University of Florida, Marianna, Florida 32446, USA
[e] Quality & Safety Assessment Research Unit, U. S. National Poultry Research Center, United States Department of Agriculture, Athens, Georgia 30605, USA

* These authors contributed equally to this work
Corresponding author: Yu-Chien Tseng
Address: Department of Agronomy, National Chiayi University, Chiayi 60004, Taiwan
Email: yct@mail.ncyu.edu.tw

# Abstract

Seed maturity in peanut (*Arachis hypogaea* L.) determines economic return to a producer because of its impact on seed weight (yield), and critically influences seed vigour and other quality characteristics. During seed development, the inner mesocarp layer of the pericarp (hull) transitions in colour from white to black as the seed matures. The maturity assessment process involves the removal of the exocarp of the hull and visually categorizing the mesocarp colours into varying colour classes from immature (white, yellow, orange) to mature (brown, and black). This visual colour classification is time consuming because the exocarp must be manually removed. In addition, the visual classification process involves human assessment of colours, which leads to large variability of colour classification from observer to observer. A more objective, digital imaging approach to peanut maturity is needed, optimally without the requirement of removal of the hull's exocarp. This study examined the use of a hyperspectral imaging (HSI) process to determine pod maturity with intact pericarps. The HSI method leveraged spectral differences between mature and immature pods within a classification algorithm to identify the mature and immature pods. Therefore, there is no need to remove the exocarp nor is there a need for subjective colour assessment in the proposed process. The results showed a consistent high classification accuracy using samples from different years and cultivars. In addition, the proposed method was capable of estimating a continuous-valued, pixel-level maturity value for individual peanut pods, allowing for a valuable tool that can be utilized in seed quality research. This new method solves issues of labour intensity and subjective error that all current methods of peanut maturity determination have.

Key Words:

classification; hyperspectral; maturity; peanut; seed quality; spectral unmixing.

## 1. Introduction

Cultivated peanut (*Arachis hypogaea* L.) is an important agronomic legume grown mainly in tropical and subtropical areas. Peanut agronomic production can be complex due to the crop's unique geocarpic reproduction, susceptibility to many fungal pathogens, and complex mechanical harvest procedure requiring separate inversion and threshing operations. Further, peanut has an indeterminate growth habit, thus having pods at various maturity levels throughout a growing season and at harvest time. It is critical for growers to accurately determine pod maturity to be able to maximize seed weight and quality, thus leading to optimal economic returns. However, the geocarpic fruit habit of peanut increases the difficulty in determining the percent of pods that are mature and ultimately to determining the optimum digging time. If digging occurs too early in development, immature pods can cause poor yield, grade, seed quality and flavour. If digging occurs too late, over-mature pods can detach from the vine during the digging or the threshing process.

Several methods have been developed for growers to predict and optimize harvest time. The most commonly accepted method is to remove the exocarp from the pericarp (hull) and categorize the inner mesocarp colour. This method was standardized by Williams and Drexler (1981) who created a Maturity Profile Board (MPB) that provided colour classification into five main colour categories (white, yellow, orange, brown, and black) and sub-categories within representing various shades within the main colour categories. Mesocarp colours of black and brown indicate mature seeds and conversely, orange, yellow, and white represent immature seeds. Based on the ratio of pods within various main and shade categories, growers can determine the optimal number of days until digging the crop (the first stage of harvest). This remains the primary method utilized by producers and researchers to date for evaluating seed maturity.

However, the MPB method has many flaws and disadvantages. First, it involves exocarp removal, often requiring "pod blasting" using a pressure washer. In this process, most white pods are blown apart and lost in the analysis due to their fragile pericarp structure and high water content. Further, this process can be extremely time consuming to reach the level of exocarp removal for accurate visual mesocarp assessment while not destroying too many of the pods in the process. Second, once the exocarp is removed, pods must be colour categorized using the human eye, a process that is extremely subjective (due to observer variability in colour categorization, lighting conditions, observer fatigue, etc.), thus introducing large error potential into the process. In addition, the process of visually categorizing pods can be very time consuming as well because it involves the individual placement of 100-150 pods within main and shade colour categories. For these reasons, the current process for determining peanut maturity is flawed and in need of a method that is objective and does not require exocarp removal.

A peanut pod maturity estimation approach was recently proposed (Bindlish, Abbott & Balota, 2017) to replace the subjective colour categorization process with a simple Nearest Neighbour classifier that classifies the median RGB colour of mesocarp of each peanut into one of ten pre-set colour classes (a.k.a ten maturity levels). In other words, it learns a simple mapping from RGB values of mesocarp images to peanut maturity. However, the time-consuming pod blasting step to remove the exocarp is still necessary in this method.

This challenge, to estimate peanut maturity on the exocarp level without pod blasting, lends itself to an evaluation of the peanut pericarp using more sophisticated imaging techniques. The current project explored using a hyperspectral imaging (HSI) process capable of detecting materials that differed in chemical composition. The colour change process in peanut involves the accumulation of tannins and other polyphenols within the mesocarp layer. It was reported that in hulls and seed coats, the tannin content increased significantly as peanut pod developed and showed a close relationship between tannin and maturity (Sanders, 1977), thus leading to a change in chemical composition within the pericarp that corresponds to seed development. If an HSI process could detect pericarp hyperspectral signatures indicative of each of the major colour classes, then peanut maturity could be determined non-destructively.

## 2. Materials and Methods

### 2.1. Theoretical Framework of the HSI

The development of the HSI process involved the collection of a high dimensional hyperspectral image data cube. This cube consisted of a stack of hundreds of two-dimensional images collected at different wavelengths across the electromagnetic spectrum, as illustrated in Figure 1. The spectral signature collected across all measured wavelengths associated with each pixel in a hyperspectral image is composed of the radiance values from each of the measured wavelengths and characterizes the chemical composition of the materials inside the corresponding field of view. In other words, pixels whose corresponding spatial area consists of materials with differing chemical compositions have different spectral signatures. By analysing the spectral signatures associated with each pixel, it is possible to identify regions with distinct material compositions.

One common method of analysis applied to hyperspectral data cubes is *spectral unmixing.* The overall goal of hyperspectral unmixing is to decompose each pixel spectrum into the collection of *endmembers* found in the pixel's field of view and their associated proportion values (Bioucas-Dias et al., 2012; Keshava & Mustard, 2002). An *endmember* is the spectral signature associated with a *pure* material found within an imaged scene. What is considered a pure material, however, is generally problem dependent and relates to the materials that are informative and relevant to the problem at hand. In our application, we considered two endmembers, one corresponding to a mature peanut and one corresponding to an immature peanut and their associated chemical compositions.

To perform hyperspectral unmixing, a *mixing model,* was assumed. The Linear Mixture Model (LMM) is a widely-used model in hyperspectral image analysis that assumes that the spectral signature associated with each pixel is a combination of endmembers weighted according to the proportion of the amount of each material found within the field-of-view. When assuming the LMM, a pixel spectrum, $\boldsymbol{x}_i$, can represented by the following:

$$\boldsymbol{x}_i = \sum_{k=1}^{M} \boldsymbol{e}_k p_{ik} + \boldsymbol{\varepsilon}_i \tag{1}$$

$$s.t. \, 0 \leq p_{ik} \leq 1, \qquad 0 \leq k \leq M \tag{2}$$

$$\sum_{k=1}^{M} p_{ik} = 1 \, , \forall i \tag{3}$$

where $\boldsymbol{\varepsilon}_i$ is the error term accounting for noise, $\boldsymbol{e}_k$ is the $k$th endmember spectrum, $p_{ik}$ is the proportion of $k$th endmember in pixel $\boldsymbol{x}_i$ and $M$ is the number of endmembers. Given this model and all of the pixels in a hyperspectral data cube, $\{\boldsymbol{x}_i\}_{i=1}^N$, the goal of the hyperspectral unmixing algorithm was to estimate the matrix of endmembers, $\{\boldsymbol{e}_k\}_{k=1}^M$, and proportion values, $\{p_{ik}\}$ which satisfy Equations 1-3. A summary of the symbols, units and abbreviations used in this work are shown in Table 4, 5 and 6, respectively.

Next, an algorithm for unmixing was chosen. Many supervised and unsupervised algorithms for hyperspectral unmixing have been developed in the literature. In general, unsupervised hyperspectral unmixing approaches estimate both the endmember spectra and proportion values without any prior knowledge of the endmember spectral signatures found within the data cube (Zare, Gader, Bchir, & Frigui, 2013; Zare, Gader, & Casella, 2013; Zou & Zare, 2017). In contrast, the endmember spectra are assumed to be known in advance and only the proportion values need to be estimated in supervised unmixing approaches (Roberts et al., 1998; Zare, Gader, Dranishnikov, & Glenn, 2013). In our approach, chose to use the mean spectral signatures of mature and immature pods as the endmembers for the mature and immature classes, respectively, and we chose to use the Fully Constrained Least Squares (FCLS) algorithm as our supervised hyperspectral unmixing approach (Heinz, Chang, & Althouse, 1999). These were chosen due to their simplicity in implementation and performance during experimentation. The FCLS algorithm estimates the proportion values, $P$, by minimizing (4) subject to the constraints in (2) and (3):

$$error = \|\boldsymbol{X} - \boldsymbol{EP}\|^2 \tag{4}$$

where $X$ is the matrix of pixel spectra, a $D$-by-$N$ matrix, represented as $\{x_i\}_{i=1}^{N}$. The $k$th endmember signature is also a 1-by-$D$ vector, represented as $e_k \in R^D$. $E$ is the matrix of endmember spectra, a $D$-by-$M$ matrix, represented as $\{e_k\}_{k=1}^{M}$. The proportion vector $p_i$ for $i$th pixel $x_i$ is a 1-by-$M$ vector, represented as $p_i \in R^M$. $P$ is the matrix of proportion values, a $M$-by-$N$ matrix, represented as $\{p_i\}_{i=1}^{N}$.

We then classified the resulting elements of the image after unmixing. The aim of hyperspectral classification is to accurately classify different regions or pixels of a hyperspectral image to various categories of interest (Camps-Valls, Marsheva, & Zhou, 2007; Chen, Nasrabadi, & Tran, 2011; El Rahman, 2016; Kuo, Li, & Yang, 2009; Melgani & Bruzzone, 2004). Peanut pixels were classified into different maturity levels by using the proportion values estimated through hyperspectral unmixing as features. We classified pods into two classes, mature vs. immature, as a first pass; then went on to separate pods into four major colour classes (yellow, orange, brown, and black).

## 2.2. Sample Preparation

To validate our HSI method, field samples were collected and tested for the accuracy of maturity assessment. Samples were collected within existing field experiments at the North Florida Research and Education Center (NFREC) near Marianna, Florida, USA (29°23″ N, 82°12″ W) in 2016 and 2017. In 2016, five cultivars were utilized including TUFRunner 511 (Tillman & Gorbet, 2017), FloRun 157, Georgia-06G (Branch, 2007), TUFRunner 297 (Tillman, 2017) and FloRun 331. In 2017, two additional cultivars, UF 08036 and FloRun 107 (Tillman & Gorbet, 2015), were added. Each cultivar had three replicated plots in the field and 200 pods were randomly selected from each plot.

Hyperspectral images were taken at the Quality & Safety Assessment Research Unit, USDA/ARS, Athens, Georgia, USA. From each field replication, 15 pods were selected from the main group of 200 pods. Ideally, the 15 pod sample would contain both mature and immature pods to provide our HSI comparison signatures; therefore, the 15 were chosen according to the shell rigidity – approximately half with flexible pericarp characteristics (high probability of being immature) and stiff pericarps (high probability of being mature). The reason to have both mature and immature pods is to provide a better ground truth dataset after destructive analysis and to provide a more accurate calibration process.

In 2016, there were 215 peanut pods (5 cultivars x 3 replications x 15 pods) and in 2017, there were 315 peanut pods (7 cultivars x 3 replications x 15 pods). After HSI acquisition, the destructive analysis initiated to obtain the ground truth data. Each peanut pod was individually blasted with a pressure washer to remove the exocarp. The colours of the mesocarp were the ground truth data from destructive analysis. They were visually classified to black, brown, orange and yellow, representing four maturity classes from mature to immature, respectively. White pods were not classified as these tend to be much smaller and do not stay intact during the blasting procedure.

## 2.3. Image acquisition and data collection

The visible-near infrared (VNIR, 400-1,000 nm) hyperspectral images were collected with a pushbroom line-scan hyperspectral imaging system configured for diffuse reflectance imaging. The system consisted of a spectrograph (ImSpector V10E, Specim, Oulu, Finland), a focal plane scanner (ITD, Stennis Space Center, MS), a 12-bit digital camera with a Peltier-cooled 1,376 x 1,040 CCD (SensiCam QE, Cooke Corp., Auburn Hills, MI), a 35mm f/4 front lens (XNP1.8/35-0901, Schneider Optics,

Hauppage, NY), and two 500W tungsten halogen lamps (Starlite QL, Photoflex, Bartlett, IL).

The camera's exposure time was 88 msec per line scan. The working distance from the lens to the scene was about 80 cm. The field of view was about 20 cm (width) x 15 cm (height). After 1x2 binning on the CCD camera, the resolution of a collected unprocessed hyperspectral image cube was 1,376 (width) x 1,000 (height, i.e. scan lines) pixels with 520 wavebands from 353-1,018 nm. The wavelength of the hyperspectral imaging system was calibrated and unchanged throughout the entire study period. Reflectance intensities were recorded for every pixel of the scanned image of the sample taken at each wavelength slice. Thus, the measured reflectance intensities were calibrated to relative reflectance values (%) with a NIST-traceable diffuse reflectance standard (99%, 10" x 10", Spectralon®, Labsphere, North Sutton, NH) as maximum reflectivity reference and a dark current measured with the lens cap on for a baseline removal.

The calibrated hyperspectral images were spectrally smoothed with the Savitzky-Golay filter (order=4, width=25). The resulting hyperspectral image cubes were spectrally de-noised and the intensity and wavelength at each image pixel and waveband slice were calibrated, resulting in dimensions of 1,376 (width) x 1,000 (height) x 467 (wavelength) in the usable wavelength range of 400–1,000 nm incremented by about 1.3 nm interval. A colour reference standard (ColorChecker Passport Photo, Xrite, Grand Rapids, MI) was also imaged together with 15 peanut pod samples per image, which were arranged to a 5 (columns) x 3 (rows) matrix form.

## 2.4. Image analysis

The approach consisted of both training and testing procedures. In the training phase, training peanut imagery was paired with a corresponding label indicating its maturity level (the mesocarp colour). Using this training imagery, the goal of the training phase was to estimate the endmember spectra for mature and immature peanuts, respectively, as well as the classification thresholds. In the testing phase, the endmembers estimated during training were used to unmix input imagery and estimate mature and immature proportion values. The proportion values were then used within a classification procedure to do final mature versus immature classification. An overview of the training and testing phases of the proposed method are shown in Figure 2. The pseudocode for the training and testing phases of the proposed method is shown in Algorithm 1 and 2.

Prior to endmember extraction, the input hyperspectral imagery was pre-processed to remove non-peanut background signatures and extract peanut pixel locations. To accomplish this, first, a colour image (Figure 3a) was extracted from the measured hyperspectral image (using the red, green and blue wavelengths). Any color calibration panels or non-uniform background materials in the imagery were manually cropped from the image. Then, the peanuts were segmented from the background material using k-means clustering where the features for each pixel were simply the RGB color values and k=2. An example segmented peanut colour image is shown in Figure 3b. As is shown, the colours of exocarp layer of pericarp are quite similar for all peanuts despite the fact that some of them are mature and some are immature. Currently, there are methods, to the best of our knowledge, able to estimate the peanut maturity based on the colour of exocarp layer of pericarp. Furthermore, given the extreme similarity in colour across maturity levels, we believe it is unlikely that colour properties of the exocarp alone can provide any discriminating information to distinguish between maturity levels.

In order to make use of hyperspectral imagery for maturity estimation, endmember spectral signatures for each class of interest (i.e., maturity level) were estimated. One endmember was associated with each class of interest. Therefore, users defined the number of endmembers by controlling the number of labels types. For instance, in order to simply classify peanuts as either immature or mature, the number of endmembers is set to two (since there are only two labels of interest). In the current study, the proposed endmember estimation strategy was spectrum averaging. Each endmember spectral signature was estimated by averaging the pixel spectra associated with each maturity level over all training peanut data. For example, for the two-class classification, the endmember spectrum was estimated as the average spectrum of all pixel spectra associated with immature peanuts or mature peanuts.

After endmember estimation, a supervised hyperspectral unmixing approach was used to unmix each peanut pixel and estimate the proportion values associated with each of the estimated endmember spectra. The estimated proportion values by unmixing represent the degree of confidence about the maturity levels of each pixel of each peanut. For instance, for two-class classification, if the proportion values for a pixel is 90% immature and 10% mature, the confidence value is 90% immature for this pixel. Proportion values of each pixel show the maturity confidence values at the individual pixel level. This provides a novel perspective on understanding how peanuts grow from immature to mature. An example of this transition is how and when the colour changes occur at different components of the peanut. To estimate a peanut-level maturity confidence, all pixel-level maturity confidences on the same peanut were aggregated.

After the maturity confidence value for each peanut was estimated, we performed the final classification. This was accomplished by thresholding the aggregated confidence value. The classification threshold was determined by selecting a threshold τ such that the misclassified rate was minimized:

$$argmin_\tau \big(FP(\tau) + FN(\tau)\big) \tag{5}$$

where FN(τ) is the number of immature peanuts that are misclassified to mature using a threshold of τ and FP(τ) is the number of mature peanuts that are misclassified to immature using a threshold of τ. The misclassified rates were calculated using different threshold values τ from 0.01 to 0.99 with a 0.01 interval. Then the optimal τ was chosen such that the misclassified rate was minimized in the training data.

## 2.5. Evaluation metrics

A confusion matrix is used to evaluate the performance of classification results. Each row of the confusion matrix denotes the samples in the actual (true) class while each column denotes the samples in a predicted class.  For instance, in Table 1, the first row and first column has a value of 116.  This indicates that are 116 actual mature samples are correctly predicted as mature. Similarly, the first row and second column has a value of six, meaning that there are six mature samples that are incorrectly predicted as immature. Thus, based on the first row, it can be interpreted that there are (116+6) actual mature samples, 116 of which are classified correctly as mature and 6 of which are classified incorrectly as immature.

There are several metrics used in this work to evaluation the performance of the method, including accuracy, precision, recall, specificity and balanced accuracy.

$$Accuracy = \frac{TP + TN}{TP + FP + TN + TN} \tag{6}$$

$$Precision = \frac{TP}{TP + FP} \tag{7}$$

$$Recall = \frac{TP}{TP + FN} \tag{8}$$

$$Specificity = \frac{TN}{TN + FP} \tag{9}$$

$$Balanced\ accuracy = \frac{Recall + Specificity}{2} \tag{10}$$

where TP denotes True Positive (i.e., immature predicted as immature), TN denotes True Negative (i.e., mature predicted as mature), FP denotes False Positive (i.e., mature predicted as immature) and FN denotes False Negative (i.e., immature predicted as mature). Balanced accuracy is included since there are an unbalanced number of samples between mature and immature peanuts.

## 3. Results and Discussion

In this study, the maturity estimation and classification experiments were trained on peanuts collected in 2016 and tested on peanuts collected in 2017, and vice versa.

### 3.1. Training on 2016 and testing on 2017 data

During training, all 2016 peanut hyperspectral images were used to generate average spectra for immature and mature peanuts. Each cultivar had similar spectral characteristics in terms of distinguishing between mature and immature peanut samples. Georgia-06G is used as an example cultivar to illustrate the spectral difference. Figure 4 shows the spectra at different wavelengths (400-1000 nm) between immature and mature peanut pods from the cultivar Georgia-06G. Our analysis focused on using the spectral responses between wavelengths of 650 and 1000 nm since the two classes have the largest difference in this range, as shown in the rectangular box in Figure 4. Then, the two average spectra were estimated using the responses in the training data from all cultivars and were regarded as the endmembers used within the FCLS unmixing approach. According to Figure 4, the average spectra suggest that the mature and immature pods in the data set can be better differentiated in the NIR spectral range than the visible spectrum. The figure supports why colour-based devices including human eyes and RGB cameras will fail to meet the application requirement. There was a significant spectral difference between mature and immature peanut pods around 970-980 nm (Figure 4). Interestingly, these bands in particular correspond to the O-H stretching second overtone of water (Bowker, Hawkins, & Zhuang, 2014; Wu & Sun, 2013), indicating that these differences may be related to the variability in water content of the pericarp and enclosed seed among maturity classes. This observation matches the general development pattern of crops where immature seeds have higher moisture than mature seeds, with moisture declining during seed maturation (Egli & Tekrony, 1997). Figure 4 also suggests that a further future investigation into the potential of the NIR spectral range is needed for exploring the potential of multispectral imaging by selecting a few key wavelengths. Another future work is to predict the amount of tannins and other polyphenols within the mesocarp layer using NIR spectroscopy or hyperspectral imaging, to determine the amount of tannins and other polyphenols can be chemically measured.

To estimate the classification threshold, the estimated endmember spectra were used to unmix the training data. Figure 5 is the estimated immature proportion map from one of the replications of peanut cultivar Georgia-06G. Only the immature proportion map is shown since each proportion value in the mature proportion map can be computed simply by subtracting the immature proportion map from one. Each proportion map shows the spatial distribution of maturity levels within a single peanut. To quantitatively evaluate the performance of unmixing results, we averaged the proportion values across each peanut sample to obtain a single confidence value, as shown in Figure 5.

Figure 6 shows the corresponding colour images of the mesocarp peanuts after pod blasting. Note that our hyperspectral images were taken on the exocarp layer of pericarp of peanuts as is shown in Figure 3(b). Then we proposed a hyperspectral unmixing and classification system to predict the maturity. The mesocarp layer pictures (Figure 6) were only used to evaluate the performance of our prediction. The maturity level ranged from yellow to black and the colour were labelled as yellow, orange, brown or black based on the ground truth estimation for the dataset. Note that the maturity ground truth based on mesocarp colour was highly subjective, especially for the peanuts in the 'transition' stage from immature to mature with orange or brown mesocarps. Based on Figure 5 and Figure 6 results, except for 6j, all peanuts show an accurate prediction of the ground truth categorization based on the proportion map. On 6c, 6e, 6h, 6l, 6m and 6o, the map showed a high value (0.711-0.909) and indicated that all peanuts were immature. On 6a, 6b, 6d, 6f, 6g, 6i, 6k, and 6n, the values were between 0.060 and 0.176, indicating some mature pod features. The only exception is 6j, and according to the Figure 6 image, it should have been labelled in the ground truth as brown instead of orange. This discrepancy between ground truth and model classification was likely due to the subjective nature of visual classification. This also shows that the proportion map of the HSI model has the potential to increase the accuracy of maturity determination.

The peanuts in the 2016 training dataset were divided into 4 groups based on their labelled mesocarp colour (black, brown, orange and yellow). Their corresponding immature confidence histograms are shown in Figure 7a, containing the results for all five peanut cultivars. The confidence histograms of yellow (immature) and black (mature) mesocarp peanuts were densely concentrated on values close to 1 and 0, respectively. However, the histograms of orange and brown mesocarp peanuts overlapped and were spread out covering a large range of confidence values, which demonstrated that there was no clear boundary between orange and brown. Each mesocarp colour showed different levels of classification accuracy: 98.2% (55/56) for black; 92.4% (61/66) for brown; 71.9% (23/32) for orange; and 97.2% (69/71) for yellow. Black and yellow colours were accurately predicted with only one and two pods misclassified, respectively. However, orange had only a 70% accuracy rate. Brown and orange colours are the transitional stages between mature and immature pod development. The change to mesocarp colour is a progressive process, oftentimes with multiple colours present together (Williams & Drexler, 1981). These transition stages are likely more difficult to distinguish and more subject to human visual classification errors.

In these results, τ was found to be 0.33 based on the immature confidence histograms shown in Figure 7a on the 2016 training data set. The classification accuracy was 92.4% (208/225) from all five cultivars over a total of 225 pods (Table 1) with classification accuracies for each cultivar listed in Table 2. All of the cultivars showed high classification accuracy. In addition, the precision, recall, specificity and balanced accuracy are also listed in Table 1. The results indicated that this new hyperspectral analysis-based classification method can be applied accurately to different peanut cultivars.

More importantly, the immature confidence histograms for 2017 testing dataset are shown in Figure 7b. Using the threshold estimated from the training set, classification accuracy is 95.2% (300/315). The overall confusion matrix is shown in Table 1. Furthermore, the precision-recall curve is shown in Figure 9. The classification accuracy for different cultivars, TUFRunner 511, FloRun 157, Georgia-06G, TUFRunner 297, FloRun 331, UF 08036 and FloRun 107, are 93.3, 97.8, 95.6, 95.6, 93.3, 97.8 and 93.3%, respectively (Table 2). Overall, there was a 93% accuracy for predicting pod color in all seven cultivars, which strengthens the advantages and application of this method to predict pod maturity. In addition, the classification accuracy was 100, 92.3, 88.6 and 97.6% for black, brown, orange, and yellow mesocarp colours, respectively (Table 3). In general, the 2017 dataset exhibited better prediction results than 2016 dataset on both cultivar and colour levels. The results on the testing dataset show similar confidence histogram and classification accuracy. Therefore, it is feasible that the endmember signatures and threshold can be estimated at one time date and then applied to data in following years.

### 3.2. Training on 2017 and testing on 2016 data

The above experiments were reversed, such that the 2017 peanut HSI dataset was used as the training dataset and the 2016 peanut HSI dataset was the testing dataset. The confidence histogram for the immature pods is shown in Figure 8a. The classification threshold $\tau$ is found to be 0.52 during the training phase, where the classification accuracy is 95.2% (300/315) (Table 1) with classification accuracies for each cultivar listed in Table 2 and mesocarp colour in Table 3. Results mimicked those from the previous experiment with similar classification performance across all cultivars and the brown and orange classes having consistently lower accuracy than black and yellow.

The estimated threshold and endmember signatures from the 2017 training dataset were directly applied to testing the 2016 dataset for classification. The immature confidence histogram is shown in Figure 8b. The overall classification accuracy on the 2016 testing dataset is 87.1% using threshold $\tau$=0.52 (Table 1). The classification accuracy for TUFRunner 511, FloRun 157, Georgia-06G, TUFRunner 297, and FloRun 331, were 86.7, 88.9, 86.7, 84.4, and 88.9%, respectively (Table 2). The classification accuracy for mesocarp colour was 89.3% (50/56), 81.82% (54/66), 71.88% (23/32) and 97.18% (69/71) for black, brown, orange and yellow, respectively (Table 3). Interestingly, with the change of $\tau$ value, the overall classification accuracy in the 2016 dataset on cultivar and colour was lower than when the 2017 dataset was the test data. This could, in part, be due to more misclassifications in the 2016 ground truth labels.

### 4. Conclusion

This study successfully developed and tested a peanut maturity classification system that involved an objective, non-destructive technique utilizing HSI. In this method, we were able to evaluate individual pod maturity with intact exocarp layers, thus eliminating the need for peanut blasting. We found that peanuts at different maturity levels have different spectral signatures when imaged with a hyperspectral sensor. Therefore, we were able to leverage spectrum difference to estimate pod maturity levels. In addition, the use of hyperspectral unmixing allowed for quantitative pixel-level maturity estimation as opposed to a subjective approach. Interestingly, the spectral differences of pods at different maturity stages appeared to be in part driven by differences in the moisture content of pods at different maturity levels. Additional benefits of the model approach show that spectra and parameters estimated from training data can be applied to the data of different years and varying peanut cultivars. All of these elements increase the logistical application value of this method, making it a practical tool in actual peanut production scenarios as well as a valuable research tool for evaluating seed at quantifiable maturity levels for performance and physiology at a basic level.

Our approach shows how applying sensor systems to novel questions in agriculture can help optimize and reduce the uncertainties in agricultural production (Gebbers & Adamchuk, 2010). This combination of sensor technology and computer science systems provides a new opportunity to improve precision agriculture as a whole (Rumpf et al., 2010). Machine learning methods have been applied to detect crop biotic stress, including the early detection of plant disease and weed detection based on spectral features (Behmann, Steinrücken, & Plümer, 2014). This study shows a novel application of these approaches that has similar agricultural utility and value to crop production.

**Acknowledgement**

This work was supported by National Peanut Board, USA; American Peanut

Shellers Association; Florida Peanut Producers Association, and National Science

Foundation, Division of Information & Intelligent Systems, USA [grant number

#1723891 "CAREER: Supervised Learning for Incomplete and Uncertain Data"].

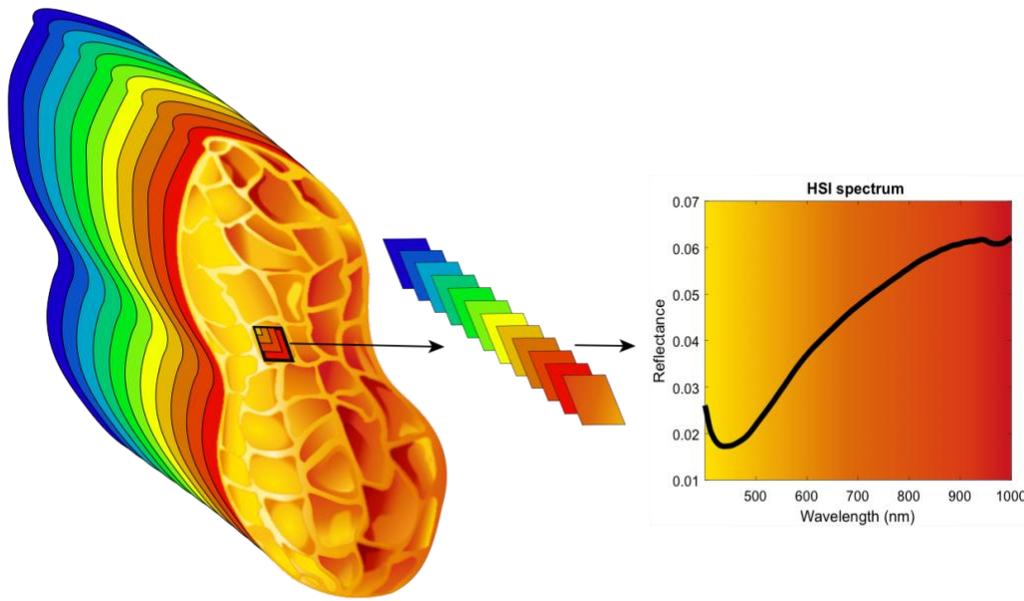

Figure 1: Illustration of a hyperspectral data cube. Left: Illustration of a hyperspectral data cube of a peanut. Each layer in this stack corresponds to the radiance values measured across the surface of the peanut at a single wavelength. Center: The collection of radiance values across all measured wavelengths associated with a single pixel (and its corresponding spatial field of view). Right: A plot of the spectral signature associated with a single pixel in the hyperspectral data cube.

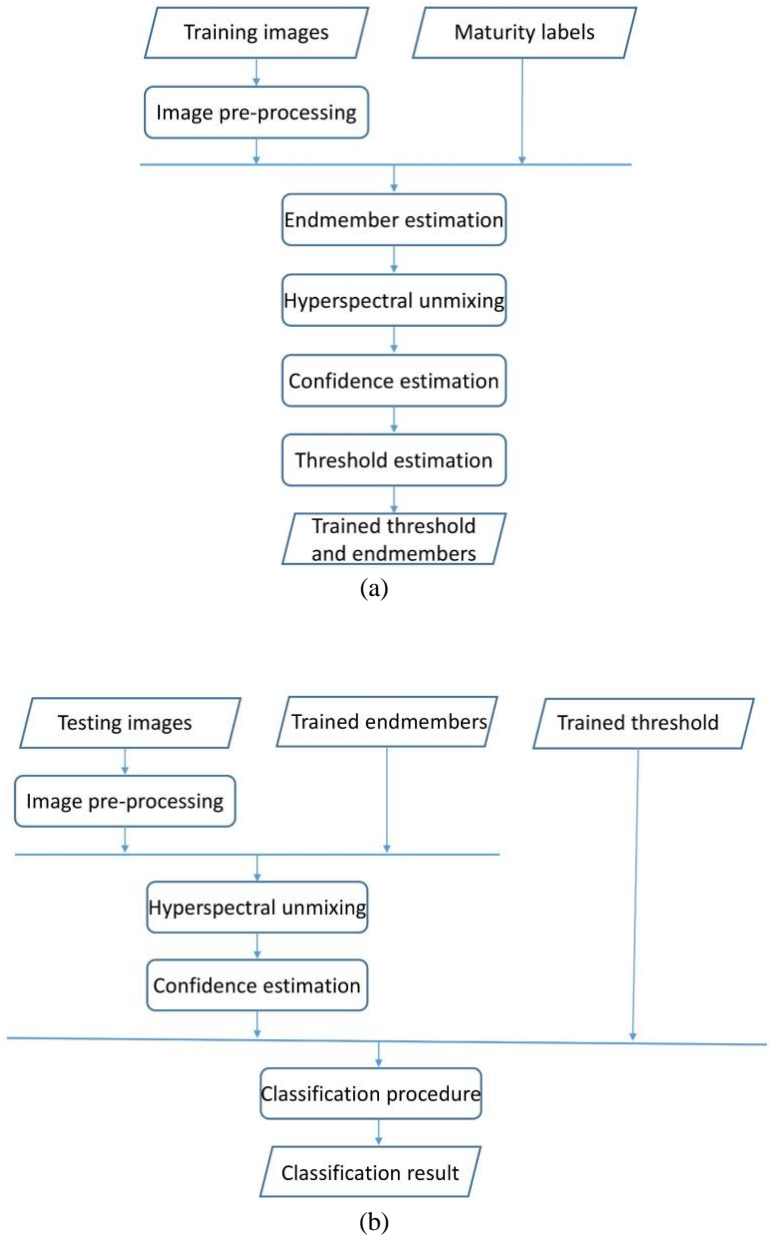

Figure 2. Flow chart of the proposed method: (a) represents the training phase; and (b) represents the testing phase.

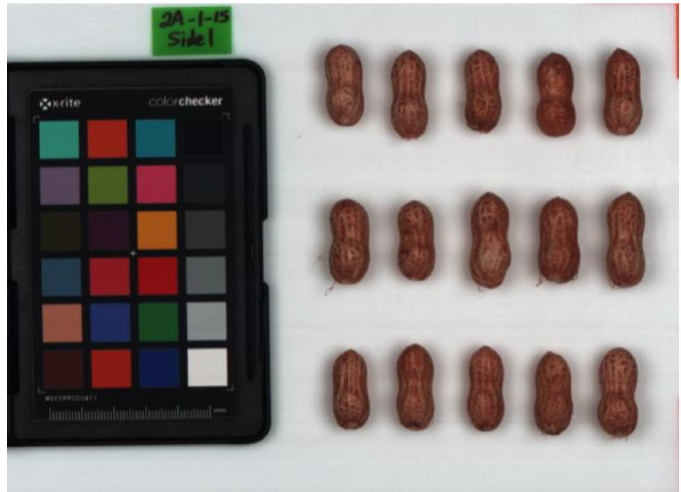

(a) RGB image

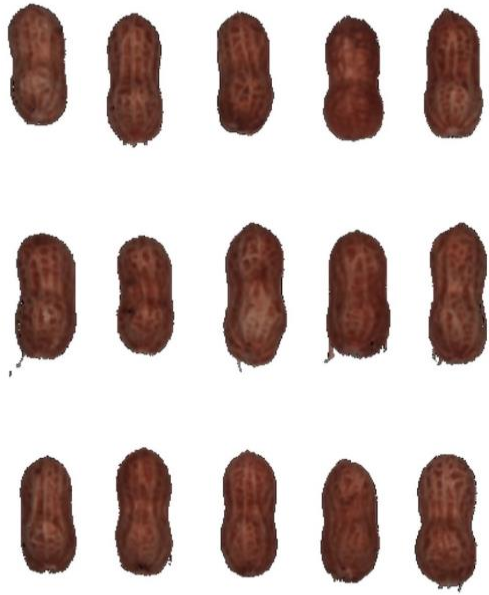

(b)

Figure 3. Illustration of peanut segmentation: (a) shows the RGB image; and (b) shows the peanut segmentation results after cropping and using K-Means.

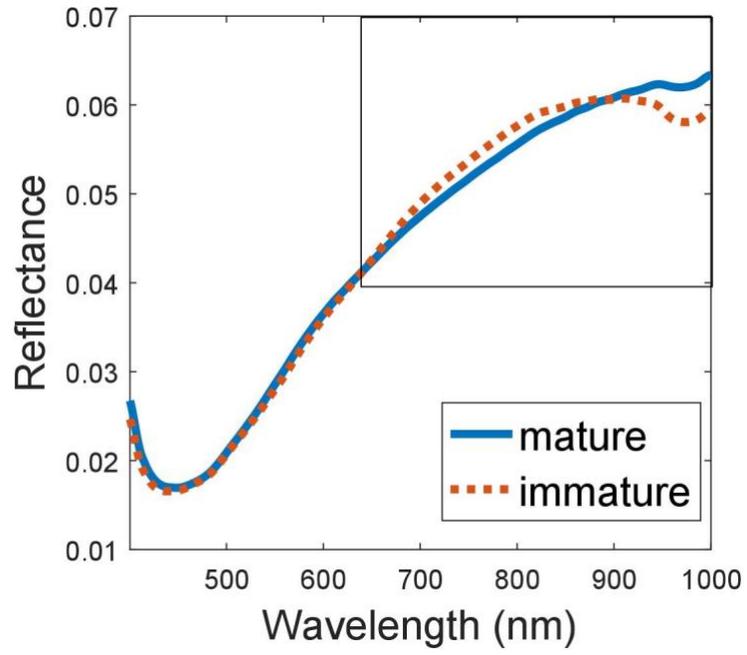

Figure 4. Average spectrum of two maturity levels, mature and immature (peanut cultivar: Georgia- 06G, replication 3).

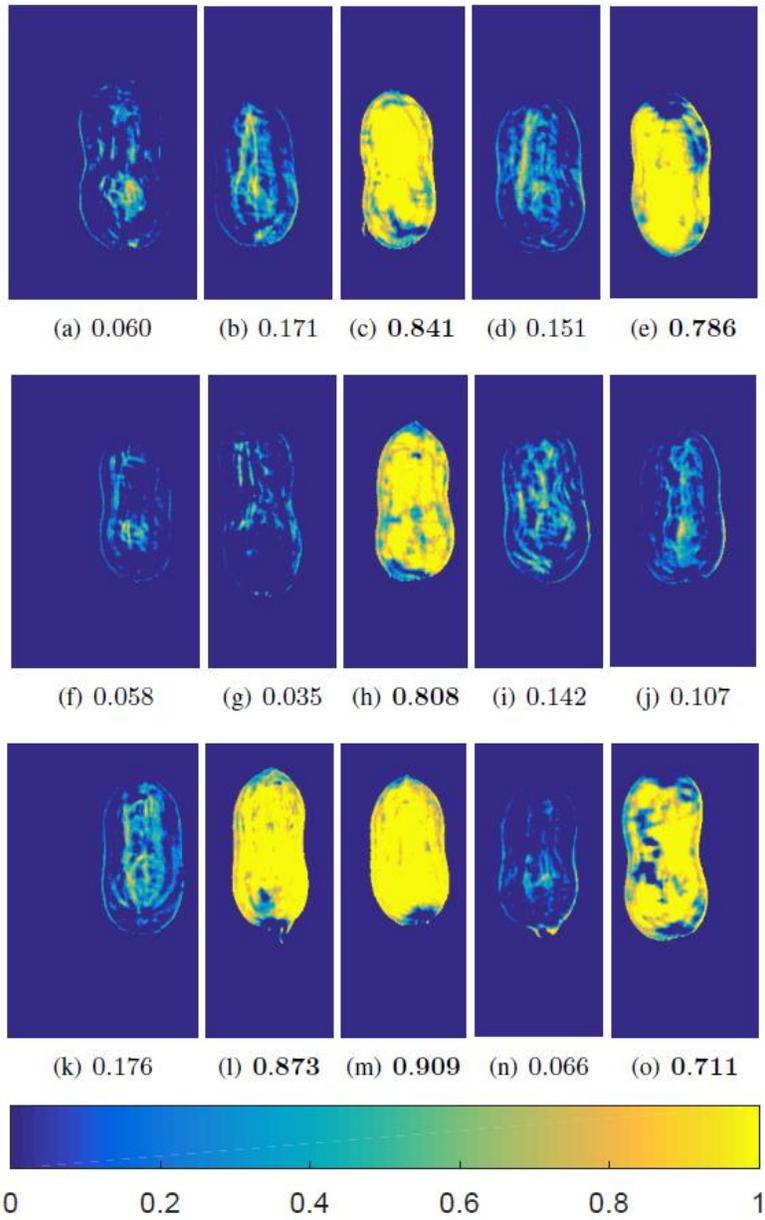

Figure 5. Estimated proportion of immature peanuts from test samples and associated immaturity confidence value (peanut cultivar Georgia-06G, replication 3).

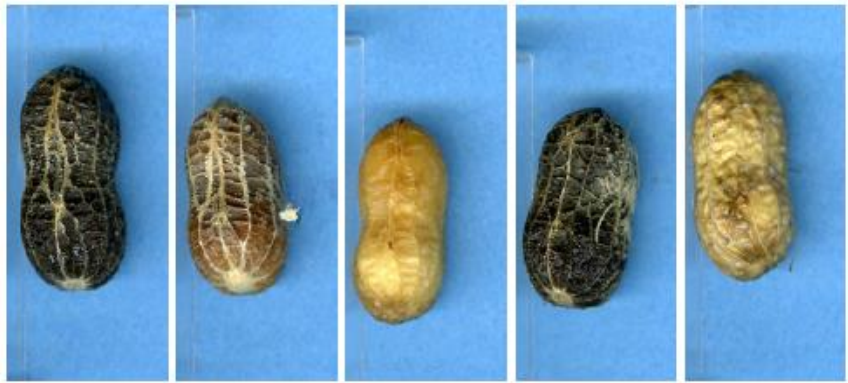

(a) Black   (b) Brown   (c) Yellow   (d) Black   (e) Yellow

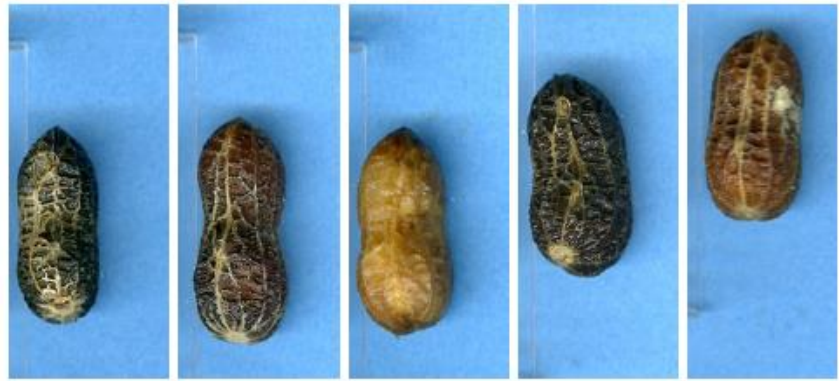

(f) Black   (g) Brown   (h) Orange   (i) Black   (j) Orange

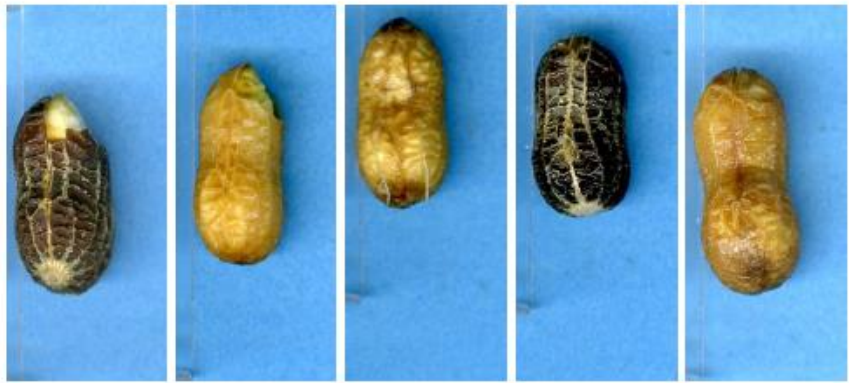

(k) Brown   (l) Yellow   (m) Yellow   (n) Brown   (o) Yellow

Figure 6. Mesocarp RGB image with associated visual classification into categories of yellow, orange (immature), brown, and black (mature) (peanut cultivar: Georgia-06G, replication 3).

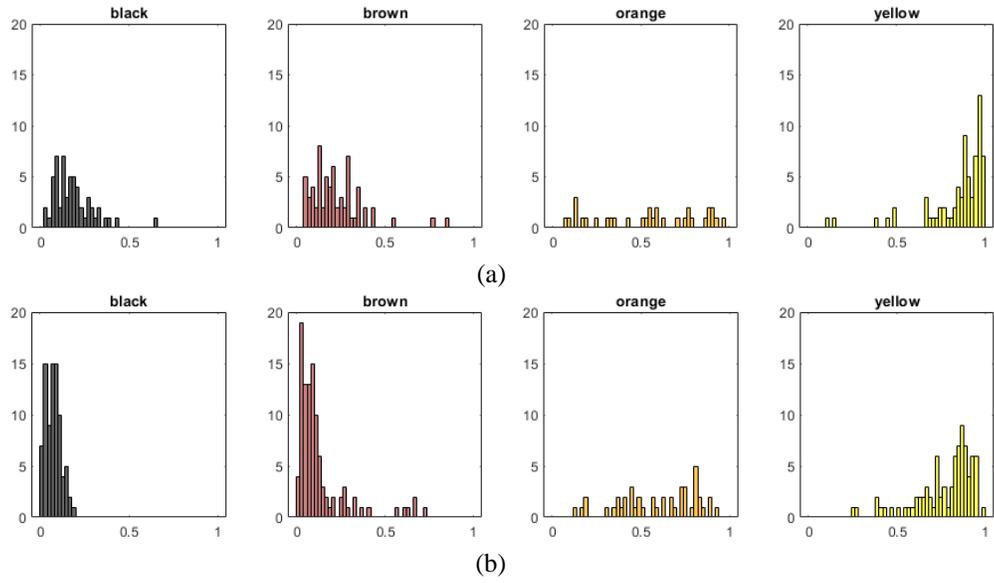

(a)

(b)

Figure 7. Confidence histograms of different peanut mesocarp colors in: (a) 2016 training dataset; and (b) 2017 testing dataset.

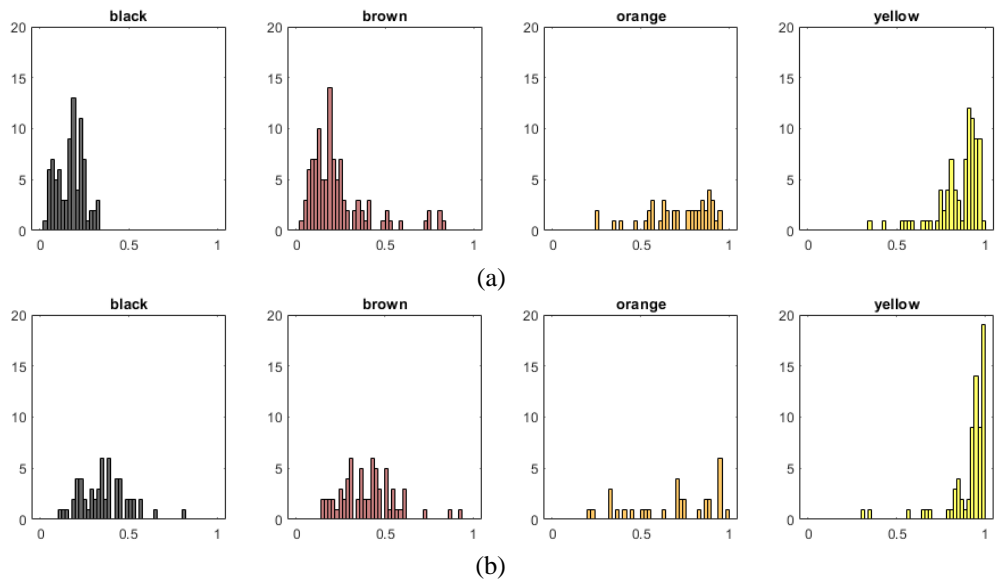

Figure 8. Confidence histograms of different peanut mesocarp colors in: (a) 2017 training dataset; and (b) 2016 testing dataset.

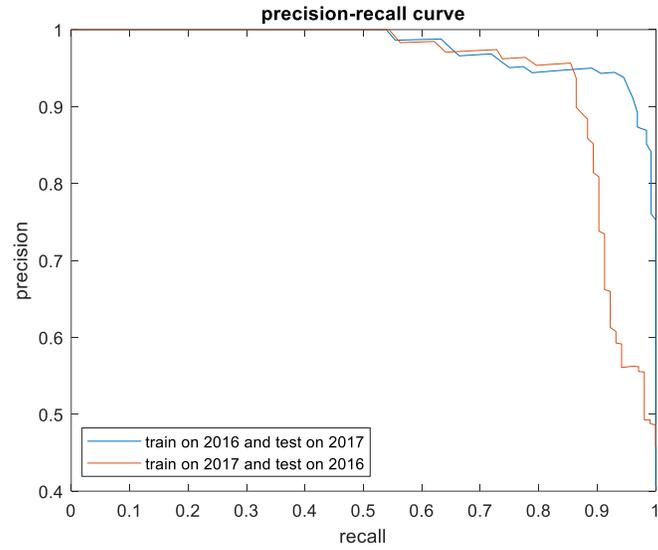

Figure 9. Precision vs recall curve for mature-vs-immature classification when: (a) training using 2016 dataset and testing on 2017 dataset; and (b) training using 2017 dataset and testing on 2016 dataset

**Algorithm 1** Peanut maturity training method

**Input**: Peanut hyperspectral images $D = \{d_1, d_2, ..., d_j\}$

**Output**: Estimated endmember spectra $E$ and threshold $\tau$

1. **Image pre-processing**: Extract RGB images from $D$. Segment out and index each peanut in the RGB images. Each peanut region hyperspectral image is denoted as $d_i$.
2. **Labeling**: Exocarp removal and visual labeling each peanut based on mesocarp color, $L_i = \{yellow, orange, brown, black\}$.
3. **Endmember estimation**: Group all peanuts based on their labels $L$ into two groups: immature and mature. Compute the average spectrum $e_m$ from mature group and $e_{im}$ from immature group. $E = \{e_m, e_{im}\}$.
4. **Hyperspectral unmixing**: Input $d_j$ and $E$ to a hyperspectral unmixing algorithm to estimate proportion values $P$ for image $j$.
5. **Confidence estimation**: Aggregate the immature proportion values $P_i$ to a confidence value $c_i$ for each peanut region.
6. **Threshold estimation**: Find a threshold value $\tau$ such that equation 7 is satisfied.
7. **Maturity classification**: $r_i = immature$ if $c_i \geq \tau$ or $r_i = mature$ if $c_i < \tau$.

---

**Algorithm 2** Peanut maturity testing method

**Input**: Peanut hyperspectral images $D = \{d_1, d_2, ..., d_j\}$, estimated endmember spectra $E$, threshold $\tau$

**Output**: Immature confidence value $c_j$ for each peanut $i$, maturity classification $r_i$ for each peanut $i$.

1. **Image pre-processing**: Extract RGB images from $D$. Segment out and index each peanut in the RGB images. Each peanut region hyperspectral image is denoted as $d_i$.
2. **Hyperspectral unmixing**: Input $d_j$ and $E$ to a hyperspectral unmixing algorithm to estimate proportion values $P$ for image $j$.
3. **Confidence estimation**: Aggregate the immature proportion values $P_i$ to a confidence value $c_i$ for each peanut region.
4. **Maturity classification**: $r_i = immature$ if $c_i \geq \tau$ or $r_i = mature$ if $c_i < \tau$.

| | 2016 (Training) | | 2017 (Testing) | | 2017 (Training) | | 2016 (Testing) | |
|---|---|---|---|---|---|---|---|---|
| | Predicted mature | Predicted immature | Predicted mature | Predicted immature | Predicted mature | Predicted immature | Predicted mature | Predicted immature |
| Actual mature | 116 | 6 | 179 | 8 | 179 | 8 | 104 | 18 |
| Actual immature | 11 | 92 | 7 | 121 | 7 | 121 | 11 | 92 |
| Accuracy rate | 92.4% | | 95.2% | | 95.2% | | 87.1% | |
| Precision | 0.939 | | 0.938 | | 0.938 | | 0.836 | |
| Recall | 0.893 | | 0.945 | | 0.945 | | 0.893 | |
| Specificity | 0.951 | | 0.957 | | 0.957 | | 0.853 | |
| Balanced acc. | 92.2% | | 95.1% | | 95.1% | | 87.3% | |

Table 1. The overall accuracy rate, precision, recall, specificity, balanced accuracy and numbers of peanut pods of different predicted and actual maturity categories for the 2016 training, 2017 testing, 2017 training and 2016 testing data.

Table 2. The classification accuracy of the 2016 training, 2017 testing, 2017 training and 2016 testing data among different peanut varieties (TUFRunner 511, FloRun 157, Georgia-06G, TUFRunner 297, FloRun 331, UF 08036 and FloRun 107).

|  | 2016 Training | 2017 Testing | 2017 Training | 2016 Testing |
|---|---|---|---|---|
| TUFRunner 511 | 91.1% | 93.3% | 93.3% | 86.7% |
| FloRun 157 | 91.1% | 97.8% | 97.8% | 88.9% |
| Georgia-06G | 95.6% | 95.6% | 95.6% | 86.7% |
| TUFRunner 297 | 88.9% | 95.6% | 95.6% | 84.4% |
| FloRun 331 | 97.8% | 93.3% | 93.3% | 88.9% |
| UF 08036 | - | 97.8% | 97.8% | - |
| FloRun 107 | - | 93.3% | 93.3% | - |

Table 3. The classification accuracy of the 2016 training, 2017 testing, 2017 training and 2016 testing data among different peanut mesocarp colors (black, brown, orange and yellow).

|        | 2016 Training | 2017 Testing | 2017 Training | 2016 Testing |
|--------|---------------|--------------|---------------|--------------|
| Black  | 98.2%         | 100%         | 100%          | 89.3%        |
| Brown  | 92.4%         | 92.3%        | 92.3%         | 81.8%        |
| Orange | 71.9%         | 88.6%        | 88.6%         | 71.8%        |
| Yellow | 97.2%         | 97.6%        | 97.6%         | 97.2%        |

Table 4. The list of symbols

| $D$ | The dimensionality of pixel spectrum, an integer in $\mathbb{R}^+$ |
|-----|--------------------------------------------------------------------|
| $\boldsymbol{E}$ | A collection of $M$ endmembers, $\boldsymbol{E} = \{\boldsymbol{e_1}, \boldsymbol{e_2}, \ldots, \boldsymbol{e_M}\}$, a matrix in $\mathbb{R}^{D \times M}$ |
| $\boldsymbol{e_k}$ | $k$-th endmember spectrum, a vector in $\mathbb{R}^D$ |
| $\boldsymbol{\varepsilon_i}$ | Noise term for $x_i$, a vector in $\mathbb{R}^D$ |
| $M$ | The number of endmembers, an integer in $\mathbb{R}^+$ |
| $N$ | The number of pixels, an integer in $\mathbb{R}^+$ |
| $\boldsymbol{P}$ | A collection of $p_{ik}$, a matrix in $\mathbb{R}^{M \times N}$ |
| $p_{ik}$ | The proportion of $e_k$ in $x_i$, a scalar in $\mathbb{R}$ of $[0,1]$ |
| $\tau$ | Classification threshold, a scalar in $\mathbb{R}$ of $[0,1]$ |
| $\boldsymbol{X}$ | A collection of $N$ pixel spectra, $\boldsymbol{X} = \{\boldsymbol{x_1}, \boldsymbol{x_2}, \ldots, \boldsymbol{x_N}\}$, a matrix in $\mathbb{R}^{D \times N}$ |
| $\boldsymbol{x_i}$ | $i$-th pixel spectrum, a vector in $\mathbb{R}^D$ |

Table 5. The list of units

| | |
|---|---|
| Nanometer | nm |
| Centimeter | cm |

Table 6. The list of abbreviations

| | |
|---|---|
| Hyperspectral imaging | HSI |
| Linear Mixture Model | LMM |
| Maturity Profile Board | MPB |
| Fully Constrained Least Square | FCLS |
| North Florida Research and Education Center | NFREC |
| Visible Near Infrared | VNIR |
| False Positive | FP |
| False Negative | FN |